\documentclass[runningheads]{llncs}

\usepackage{iciap}

\usepackage{iciapabbrv}

\usepackage{tcolorbox} 
\usepackage{graphicx}

\usepackage{amsmath}
\usepackage{amssymb}
\usepackage{bm}
\usepackage{setspace}

\usepackage{booktabs}
\usepackage{multirow}
\usepackage{makecell}
\usepackage{rotating}
\usepackage{graphicx}
\usepackage{colortbl} 
\usepackage{pgf}      
\usepackage{caption}

\usepackage{comment}

\usepackage[accsupp]{axessibility}  

\usepackage{hyperref}

\usepackage{orcidlink}

\begin{document}

\title{A Second-Order Perspective on Pruning at Initialization and Knowledge Transfer}

\titlerunning{A Second-Order Perspective on PaI and Knowledge Transfer}

\author{Leonardo Iurada$^*$\orcidlink{0009-0009-5637-9850} \and
Beatrice Occhiena \and
Tatiana Tommasi\orcidlink{0000-0001-8229-7159}}

\authorrunning{L.~Iurada et al.}

\institute{Politecnico di Torino, Italy\\
$^*$Corresponding Author: \email{leonardo.iurada@polito.it}}

\maketitle

\begin{abstract}
    The widespread availability of pre-trained vision models has enabled numerous deep learning applications through their transferable representations.
    However, their computational and storage costs often limit practical deployment. Pruning-at-Initialization has emerged as a promising approach to compress models before training, enabling efficient task-specific adaptation. While conventional wisdom suggests that effective pruning requires task-specific data, this creates a challenge when downstream tasks are unknown in advance. 
    In this paper, we investigate how data influences the pruning of pre-trained vision models. 
    Surprisingly, pruning on one task retains the model's zero-shot performance also on unseen tasks. Furthermore, fine-tuning these pruned models not only improves performance on original seen tasks but can recover held-out tasks' performance. We attribute this phenomenon to the favorable loss landscapes induced by extensive pre-training on large-scale datasets.
    
\end{abstract}
\vspace{-6pt}

\vspace{-12pt}
\vspace{-3mm}
\section{Introduction}
The surge in digital data availability has fueled the development of large-scale \emph{foundation} models which are now indispensable tools across a broad spectrum of applications. Through extensive training, these models learn rich representations with impressive generalization capabilities, achieving strong performance on a variety of downstream tasks with minimal fine-tuning, and often even in zero-shot settings. Nonetheless, their sheer size and complexity continue to pose significant challenges regarding computational and storage resources, leading to a growing interest in model compression strategies to enhance efficiency. 

\smallskip\noindent
The vast majority of literature on model \emph{pruning} focuses on reducing the inference latency for real-time deployment, although there is also a growing interest in approaches to eliminate unnecessary weights before the network’s training. \emph{Pruning-at-initialization} (PaI) identifies sparse sub-networks based on network topology and theoretically estimated learning dynamics. It is effective both when the initialization is random (learning from scratch) and when starting from a pre-trained model. Indeed, compressing networks before fine-tuning has the potential to reduce the cost of model adaptation, thereby improving accessibility to foundation models especially for resource-constrained devices. However it comes with the unique challenge of preserving the pre-trained model's original broad generalization regardless of the task-specific data used to guide the pruning process.  

\smallskip\noindent
We dedicate this work to an in-depth investigation of whether and to what extent it is possible to directly compress pre-trained vision models without compromising their transferability. The study is particularly relevant for scenarios in which the downstream application is not known in advance. In such cases, pruning is guided by data from a \emph{source} task, while performance is evaluated on a different but related \emph{target} task. This setting enables a rigorous assessment of knowledge retention and addresses a significant gap in the current PaI literature. 

\noindent\textbf{Our contributions can be summarized as follows:}

\noindent\textbf{1.} We set up a controlled environment and run an extensive analysis on the transfer performance of large-scale pre-trained pruned models, considering both convolutional neural networks and vision transformers, encompassing all major PaI techniques for unstructured pruning. 

\noindent\textbf{2.} We offer theoretical insights about the transfer learning potential of pruned models by analyzing the loss landscape and its curvature, studying the effects of weight perturbation induced during pruning and subsequent fine-tuning. 

\noindent\textbf{3.} Notably, we observe distinct trends between architectures but minimal variation across their pre-training configurations with a general conclusion about the possibility of pruning on a task and retaining performance on other tasks with distinct labeling functions, without requiring any data from the latter.
\vspace{-3mm}

\section{Related Works}

\noindent\textbf{Neural network pruning.} The challenge of reducing neural network parameters while preserving performance has been studied since the 1980s \cite{lecun1989optimal, hassibi1992second}.
The \emph{Lottery Ticket Hypothesis} \cite{frankle2018lottery} conjectured that within large networks there exist smaller sub-networks (``winning tickets'') that, if correctly identified and trained, can match the performance of the original full model. 
The Iterative Magnitude Pruning (IMP) algorithm discovers these sub-networks through alternating training and pruning phases, with high computational costs.
This led to the development of more efficient paradigms such as \emph{Pruning-at-Initialization} (PaI) which searches for ``winning tickets'' before any training, favoring a more efficient downstream task adaptation. PaI approaches fall into two categories. \emph{Data-agnostic} methods evaluate the parameters' importance without considering training data. Notable examples include SynFlow \cite{tanaka2020synflow}, which assesses parameter relevance through synaptic saliency, NTK-SAP \cite{ntksap2023wang}, which leverages Neural Tangent Kernel (NTK) theory for more precise training dynamics estimation and PHEW \cite{patil2021phew}, which scores parameter importance using biased random walks in a single-pass. 
\emph{Data-driven} methods incorporate training data and architecture information to guide pruning. SNIP \cite{lee2018snip} scores parameters based on their impact on the initial loss, while GraSP \cite{grasp2020wang} uses gradient norm changes to identify important parameters. Furthermore, PX \cite{iurada2024finding} improved on NTK-based data-agnostic methods by incorporating task data information.

\smallskip
\noindent\textbf{Transfer Learning in the Context of Pruning.} Pruning is often applied to obtain highly task-specific models optimized for a given task. However, in many real-world scenarios, only limited or even no task-specific data may be available during pruning. While this challenge could be addressed by \emph{data-agnostic} PaI methods on randomly initialized networks, it has been shown that such approaches fail in the context of pre-trained vision models \cite{iurada2024finding}.
Recent literature has explored combining pruning with transfer learning to address this issue.
In the NLP domain, \cite{dery2023transfer} applied structured pruning and re-training to BERT \cite{devlin2019bert}, leveraging transfer learning under limited task-specific data. Although they operate in a low-data regime, their approach still assumes access to transfer task data. Similarly, \cite{gordon2020compressing} compresses BERT during pre-training and analyzes the effects on downstream transfer. They find that pruning once during pre-training is sufficient, obviating the need for task-specific pruning for each downstream task.
In the vision domain, \cite{spadaro2023shannon} demonstrates a correlation between the number of unpruned parameters in the network and the entropy of its learned representation. The authors of \cite{liu2021transtailor} propose an alternative strategy by jointly tuning the weights and structure of pre-trained vision models to better align with the transfer task. However, their method requires access to both the source and transfer task data during pruning and re-training.

\smallskip
\noindent In this work, we focus on vision models extensively pre-trained on large-scale datasets covering a wide range of tasks. We apply PaI methods, simulating in a controlled setting the scenario where downstream tasks are unknown in advance. Specifically, we use data from a single \emph{source} task for pruning and re-training, then evaluate the resulting models on multiple held-out \emph{target} tasks. By analyzing how feature quality is affected by pruning, we aim to understand the role of data and to which extent compressed representations allow for transfer learning.

\vspace{-4mm}
\section{Transfer Learning of Pruned Pre-trained Models}

\noindent\textbf{Background \& problem formulation.} Consider a neural network $f: \mathbb{R}^d \times \mathbb{R}^{p} \rightarrow \mathbb{R}^C$ taking inputs $\bm{x} \in \mathbb{R}^d$ and parametrized by weights $\bm{\theta} = \{ \bm{W},\bm{\xi} \} \subseteq \mathbb{R}^{p}$. 
We assume our $f (\bm{x}, \bm{\theta}) = \varphi(\bm{x}, \bm{W})^\top \bm{\xi}$ to be composed of a feature extractor $\varphi: \mathbb{R}^d \times \{\bm{W}\} \rightarrow \mathbb{R}^h$, mapping inputs to $h$-dimensional feature vectors and a one-versus-rest linear classifier with weight matrix $\bm{\xi} \in \mathbb{R}^{h \times C}$, on which we take the maximal value (over the $C$ classes) as the final model's prediction.

\smallskip
\noindent For simplicity, we assume two multi-class \emph{classification} tasks with their overall data support $\mathcal{D} = \mathcal{D}_s \cup \mathcal{D}_t$ composed by two disjoint sets (\ie $\mathcal{D}_s \cap \mathcal{D}_t = \varnothing$). One defines the \emph{source} task $\mathcal{D}_s = \{(\bm{x}_i^s, f_s^\star(\bm{x}_i^s)=y_i^s )\}_{i \leq n_s}$ and the other is the \emph{transfer} task $\mathcal{D}_t = \{(\bm{x}_i^t, f_t^\star(\bm{x}_i^t)=y_i^t)\}_{i \leq n_t}$. Here $n_s$, $n_t$ indicate the respective data cardinality for the two sets and $f_s^\star: \mathcal{D}_s \rightarrow \mathcal{Y}_s$ is the target function that maps inputs to the ground truth labels. By analogy we define $f_t^\star: \mathcal{D}_t \rightarrow \mathcal{Y}_t$ for the transfer task.

\noindent
Let us denote with $\bm{\theta}_0$ the parameters obtained via extensive pre-training on a large corpus of data. The problem of \emph{unstructured} neural network pruning \cite{lee2018snip} can be formalized as finding a binary mask $\bm{c} \in \{0,1\}^p$ that is optimal for
\begin{equation}\label{eq:pruning_problem}
    \min_{\bm{c}} \mathbb{E}_{(\bm{x}, y) \sim \mathcal{D}}[L(f(\bm{x}, \mathcal{A}(\bm{\theta}_0, \bm{c}) \odot \bm{c}), y)] ~~~\text{s.t.}~ \bm{c} \in \{0,1\}^p, ~~\frac{\|\bm{c}\|_0}{p} \leq k= 1 - q
\end{equation}
where $L$ is the loss function (\eg MSE), $q$ is the desired sparsity ratio, $\mathcal{A}$ is an optimization algorithm (\eg SGD) that takes as input the mask and the initial weights and returns the pruned weights at convergence, and $\odot$ denotes the element-wise product.
In the following, we will focus on the problem in \cref{eq:pruning_problem} by assuming to only have access to the \emph{source} task data $\mathcal{D}_s$.

\vspace{-3mm}
\subsection{A Second-Order Perspective On Pruning Pre-trained Models}\label{sec:predictions}

Solving the problem in \cref{eq:pruning_problem} poses significant challenges. However, assuming $\bm{\theta} = \bm{\theta}_0$ is a point of local minimum for the empirical risk $L(\mathcal{D}, \bm{\theta})$, the second-order approximation of the loss around $\bm{\theta}_0$ allows to derive a tractable closed-form solution \cite{hassibi1992second}. Specifically, its second-order Taylor expansion around $\bm{\theta}_0$ is
\begin{equation}\label{eq:second_order_taylor}
    L(\mathcal{D}, \bm{\theta}) = L(\mathcal{D},\bm{\theta}_0) + (\bm{\theta} - \bm{\theta}_0)^\top \textbf{H} (\bm{\theta} - \bm{\theta}_0) + \mathcal{O}(\|\bm{\theta} - \bm{\theta}_0\|^3) ~,
\end{equation}
with $\textbf{H} \triangleq \nabla_{\bm{\theta}}^2 L(\mathcal{D},\bm{\theta}_0)$ identifying the Hessian around $\bm{\theta}_0$. Given that the Hessian is positive semi-definite for $\bm{\theta} = \bm{\theta}_0$ and considering the quadratic nature of \cref{eq:second_order_taylor}, such expansion defines a locally convex basin for $L(\mathcal{D},\bm{\theta})$. Thus, as done by \cite{hassibi1992second}, we can derive the optimal parameter to remove by deriving in closed-form the incurring increase in loss $\varepsilon_{[j]}$ of removing the $j$-th weight $\bm{\theta}_{[j]}$ and the optimal update $\bm{\delta}_{[j]}$ to adjust for its removal,
\begin{equation}\label{eq:obs_score+update}
    \varepsilon_{[j]} = \frac{\bm{\theta}_{[j]}^2}{[\textbf{H}^{-1}]_{[jj]}}, ~~~\bm{\delta}_{[j]} = -\frac{\bm{\theta}_{[j]}}{[\textbf{H}^{-1}]_{[jj]}} \cdot \textbf{H}^{-1}_{[:,j]} ~.
\end{equation}
Although optimal, these quantities assume access to the whole $\mathcal{D}$, rather than only $\mathcal{D}_s$ (as our assumption). However, recent insights on the regularities induced by large-scale pre-training \cite{iurada2025efficient} reveal that, if it is true that $\mathcal{D}_s,\mathcal{D}_t$ are similar to data seen during pre-training, then, we can make the following predictions (which we will empirically test via controlled experiments in \cref{sec:experiments}).
\begin{tcolorbox}[colframe=black, width=\textwidth, boxrule=0.5mm] 
\noindent\textbf{P1.} We expect that \emph{parameters not important for $\mathcal{D}_s$ are also not important for $\mathcal{D}_t$}. Thus, guiding pruning using only $\mathcal{D}_s$ is enough to compress models in an approximately optimal way.

\smallskip
\noindent\textbf{P2.} If \textbf{P1} is verified and re-training after pruning can recover performance on $\mathcal{D}_s$ (\eg via SGD or using $\bm\delta_{[j]}$), then it will also recover performance on $\mathcal{D}_t$, as the updated pruned weights will also have a lower empirical risk on the \emph{target} task, given they will be closer to $\bm\theta_0$ (which are optimal for all tasks) in the Riemann manifold \cite{lee2006riemannian} induced by $\textbf{H}$.
\end{tcolorbox}
\begin{figure}[t!]
    \centering
    \vspace{-3mm}
    \includegraphics[width=0.30\linewidth]{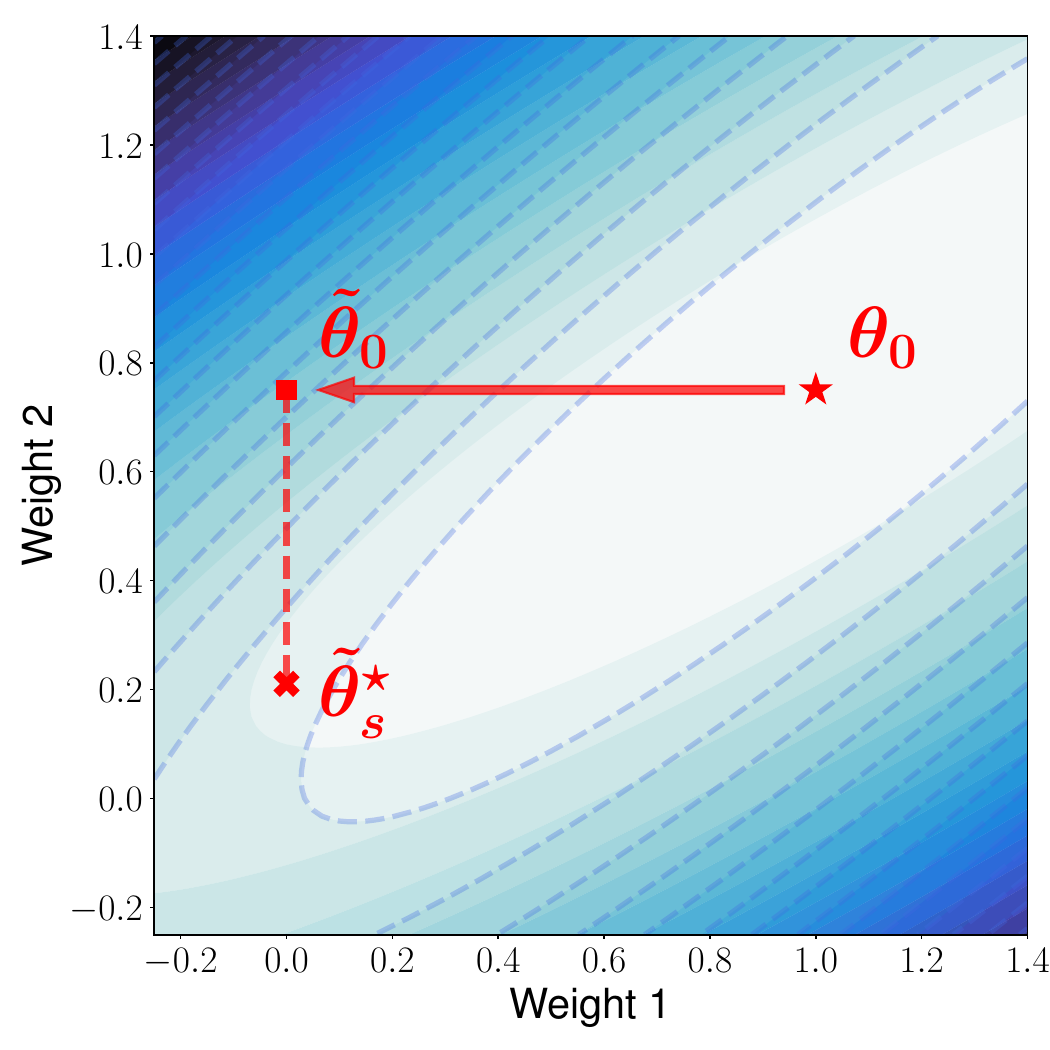}
    \includegraphics[width=0.30\linewidth]{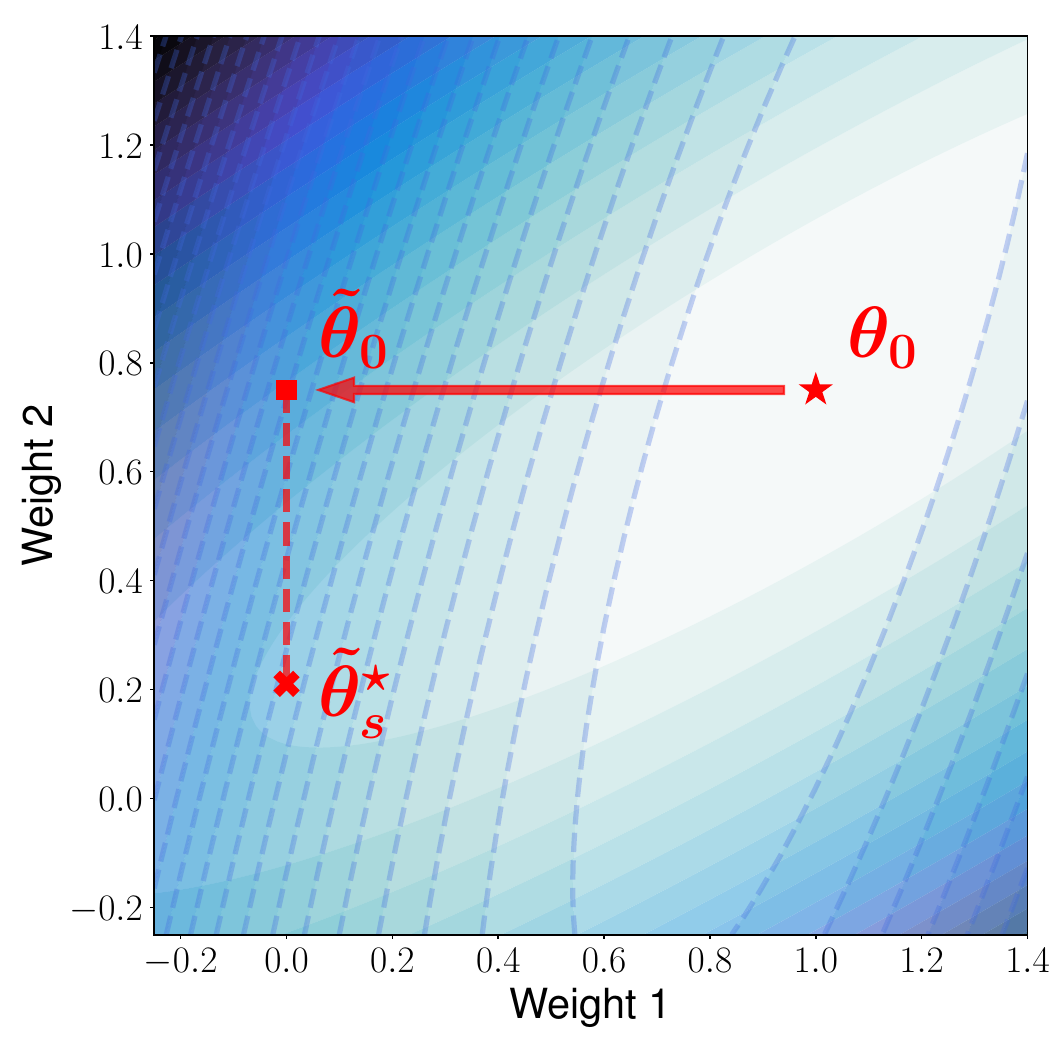}
    
    \vspace{-2.5mm}\caption{\small\textbf{Toy loss landscapes of pruned models.} We visualize the loss landscapes of toy models with two parameters. Dashed blue lines indicate the \emph{transfer} task’s loss, not observable during pruning and re-training. Blue shades represent the \emph{source} task’s loss. Both losses are modeled as quadratic functions measuring squared distance from the shared minimum $\bm\theta_0$, with rotated principal directions to simulate varying task alignment. $\tilde{\bm\theta}_0$ indicates the pruned weights, while $\tilde{\bm\theta}_s^\star$ the pruned and re-trained weights, both for the \emph{source} task.
    The red arrow indicates the weight perturbation, resulting from pruning a weight, while the red dashed line the re-training on the \emph{source} task.
    \textbf{(Left)} Aligned tasks.
    \textbf{(Right)} Misaligned tasks.
    Light colors indicate lower loss.
    }
    \label{fig:conjectures}
    \vspace{-3mm}
\end{figure}

\smallskip
\noindent We visualize our two conjectures \textbf{P1} and \textbf{P2} in \cref{fig:conjectures}. If the \emph{source} and \emph{transfer} tasks' loss landscapes are aligned, pruning on one or the other is the same. Furthermore, re-training, after pruning, only on \emph{source} moves the pruned weights back again in a point of lower loss also for the \emph{transfer} task. Conversely, if the two tasks have misaligned landscapes (in an extreme case, orthogonal) then pruning and re-training will not help with transferability as the pruned solution moves orthogonal to the contour lines of the \emph{transfer} task.

\smallskip
\noindent In the following section we will cast the problem of \cref{eq:pruning_problem} as a \emph{Pruning-at-Initialization} problem \cite{lee2018snip, grasp2020wang, tanaka2020synflow, ntksap2023wang, iurada2024finding} and introduce the main methodology we will adopt to assess our predictions \textbf{P1} and \textbf{P2}.

\vspace{-4mm}
\subsection{Adopted Pruning-at-Initialization Pipeline}\label{sec:pai_pipeline}

Given a task $\mathcal{D}_i$ on which we want to compress the weights $\bm\theta_0$, the general pipeline of \emph{Pruning-at-Initialization} (PaI) methods to tackle the problem of \cref{eq:pruning_problem} starts by assigning each weight $j \in 1, \dots, p$ a saliency score of the form
\begin{equation}
    S({\bm{\theta}_0}_{[j]}) = \frac{\partial F}{\partial {\bm{\theta}}_{[j]}} \cdot {\bm{\theta}_0}_{[j]}
\end{equation}
where $F$ is a function of the weights evaluated at $\bm\theta = \bm{\theta}_0$, capturing some property of the network. For instance, in {SNIP} \cite{lee2018snip} the saliency is the empirical risk at-initialization: $F=\mathbb{E}_{(\bm{x},y)\sim\mathcal{D}_i}[L(f(\bm{x}, \bm{\theta}_0 \odot \bm{c}), y)]$. Thus, it assigns to each parameter a score which reflects how the initial loss would change when removing that specific parameter from the network.
Once the saliency scores are assigned to each weight, we keep the top-$k$ weights by setting the masks with the highest saliency scores to one (\ie $\forall j \in \text{top}_k(S), ~\bm{c}_{[j]} = 1$), while the other weights will be pruned (\ie $\forall j \notin \text{top}_k(S), ~\bm{c}_{[j]} = 0$). Then, the pruned model $f(\cdot, \bm{\theta}_0 \odot \bm{c})$ is re-trained on $\mathcal{D}_i$ to recover its original performance.

\smallskip
\noindent Setting $S({\bm{\theta}_0}_{[j]}) = \hat{\varepsilon}_{[j]}$ as in \cref{eq:obs_score+update} provides an optimal score. However, computing this quantity requires to calculate outer products between gradients to estimate the Hessian which incurs a complexity of approximately $\mathcal{O}(p^2)$ in both time and memory.
Moreover, computing matrix inverses further raises the complexity to $\mathcal{O}(p^3)$, with the number of parameters $p$ ranging from millions to billions. 
To bypass this, standard practice is to adopt Hessian approximations \cite{lecun1989optimal, hassibi1992second, sun2023wanda, frantar2023sparsegpt} that incur in a lower cost by sacrificing some precision. Specifically,

\smallskip
\noindent\textbf{Isotropic.} By approximating the Hessian as the identity matrix $\textbf{H} = I$, the score is only based on the weight magnitude. Thus, $\hat\varepsilon_{[j]}^{mag} = \bm\theta_{[j]}^2$. This is the cheapest approximation possible for the Hessian, which assumes the curvature of the loss landscape to be isotropic around $\bm\theta_0$.

\smallskip
\noindent\textbf{Diagonal}\cite{lecun1989optimal, sun2023wanda}\textbf{.} This approximation assumes each parameter of the model as independent. Computing only the diagonal elements of the Hessian incurs in a cost of approximately $\mathcal{O}(p)$ in both time and memory and can be done efficiently by computing per-example gradients as $\textbf{H}_{[jj]} = \mathbb{E}_{(\bm{x},y) \sim \mathcal{D}_i} [\left(\nabla_{\bm\theta_{[j]}} L((\bm{x},y),\bm\theta_0)\right)^2]$.
Additionally, the inverse of a diagonal matrix is the reciprocal of its diagonal elements, bypassing the $\mathcal{O}(p^3)$ cost of computing the inverse. Thus, under this approximation, $\hat\varepsilon_{[j]}^{diag} = \bm\theta_{[j]}^2 \cdot \textbf{H}_{[jj]}$.

\smallskip  
\noindent\textbf{Block-Diagonal}\cite{hassibi1992second,frantar2023sparsegpt}\textbf{.} By grouping parameters layer-wise into blocks, this approximation assumes independence between layers, discarding cross-layer curvature terms. For layer $l$ with parameters $\bm\theta_{[l]} \in \mathbb{R}^{p_l}$, its Hessian block $\textbf{H}_{[l]} \in \mathbb{R}^{p_l \times p_l}$ is approximated as $\textbf{H}_{[l]} \approx \textbf{A}_{[l]} \otimes \textbf{B}_{[l]}$, where $\otimes$ denotes the Kronecker product, $\textbf{A}_{[l]} \in \mathbb{R}^{m_l \times m_l}$ and $\textbf{B}_{[l]} \in \mathbb{R}^{h_l \times h_l}$ are small symmetric matrices derived from the expectation of gradient outer products over the data distribution $\mathcal{D}_i$ \cite{martens2015optimizing, eschenhagen2023kronecker}. Specifically, for weight gradients $\nabla_{\bm\theta_{[l]}}L$ reshaped into a $m_l \times h_l$ matrix (with $m_l \cdot h_l = p_l$), from the chain rule $\textbf{A}_{[l]}$ and $\textbf{B}_{[l]}$ correspond to the covariances of the layer's input activations and gradient signals, respectively:  
\begin{equation}
    \textbf{A}_{[l]} = \mathbb{E}_{(\bm{x},y) \sim \mathcal{D}_i}\left[\bm{a}_{[l-1]} \cdot  \bm{a}_{[l-1]}^\top\right], \quad \textbf{B}_{[l]} = \mathbb{E}_{(\bm{x},y) \sim \mathcal{D}_i}\left[\nabla_{\bm{z}_{[l]}}L \cdot \nabla_{\bm{z}_{[l]}}L^\top\right],
\end{equation}  
where $\bm{a}_{[l-1]} \in \mathbb{R}^{m_l}$ is the input to layer $l$ and $\nabla_{\bm{z}_{[l]}}L \in \mathbb{R}^{h_l}$ the gradient of the loss \wrt layer $l$'s pre-activation output $\bm{z}_{[l]}$. The Kronecker structure reduces memory from $\mathcal{O}(p_l^2) = \mathcal{O}((m_l \cdot h_l)^2)$ to $\mathcal{O}(m_l^2 + h_l^2)$ for an $m_l \times h_l$ weight matrix, and enables efficient inversion via $\textbf{H}_{[l]}^{-1} \approx \textbf{A}_{[l]}^{-1} \otimes \textbf{B}_{[l]}^{-1}$, replacing the $\mathcal{O}(p_l^3)$ cost of direct inversion with a lower cost of $\mathcal{O}(m_l^3 + h_l^3)$. This allows layer-wise computation of $\hat\varepsilon_{[j]}^{block}$ using \cref{eq:obs_score+update}, with per-parameter scores derived from local blocks rather than the full Hessian.

\smallskip
\noindent Finally, as we only have access to $\mathcal{D}_s$, we restrict the computation of the Hessian on the data of the \emph{source} task.
Given these approximations, the resulting closed-form update $\bm{\delta}_{[j]}$ may no longer be optimal. To address this, we align with the PaI framework and, instead, we update the remaining weights using SGD.

\begin{figure}[t!]
    \centering
    \vspace{-3mm}

    \includegraphics[width=0.30\linewidth]{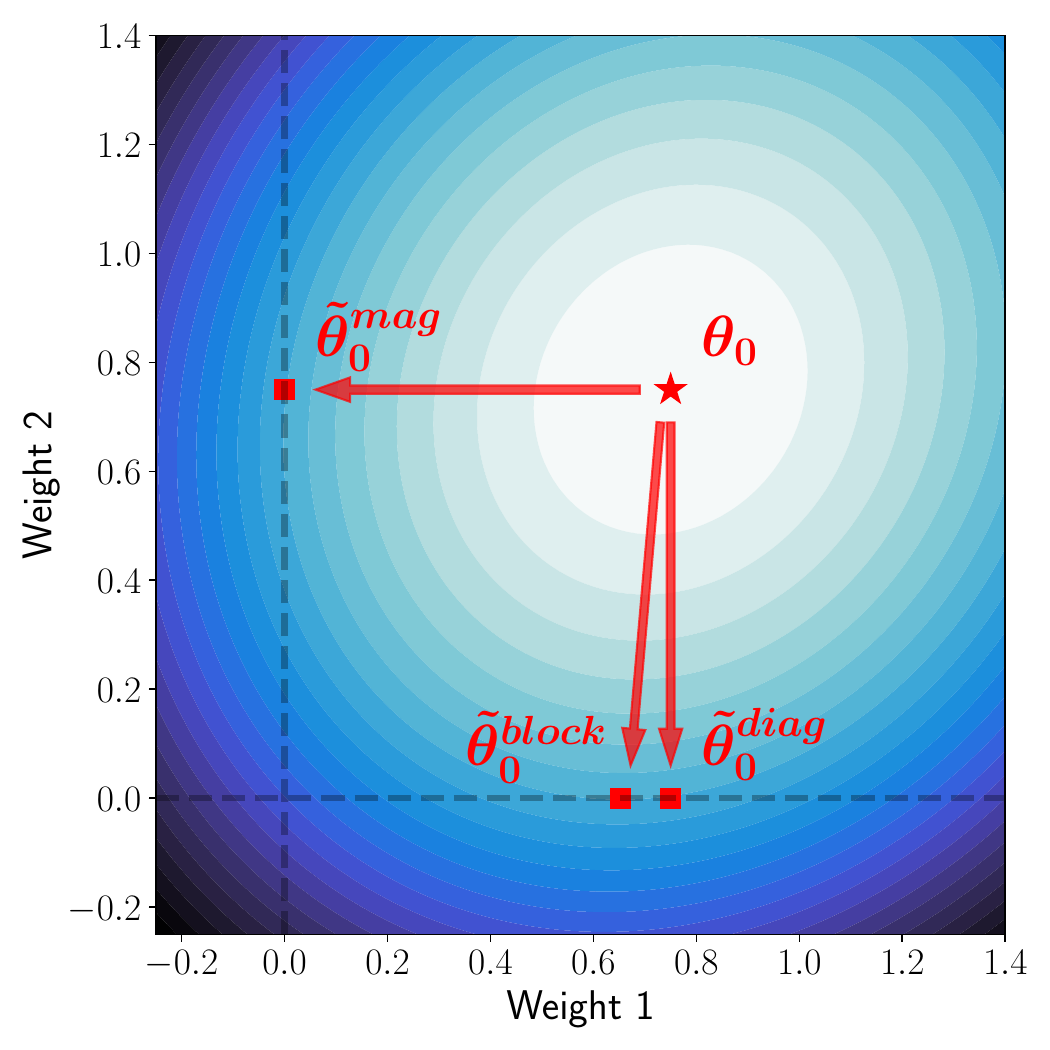}
    \includegraphics[width=0.30\linewidth]{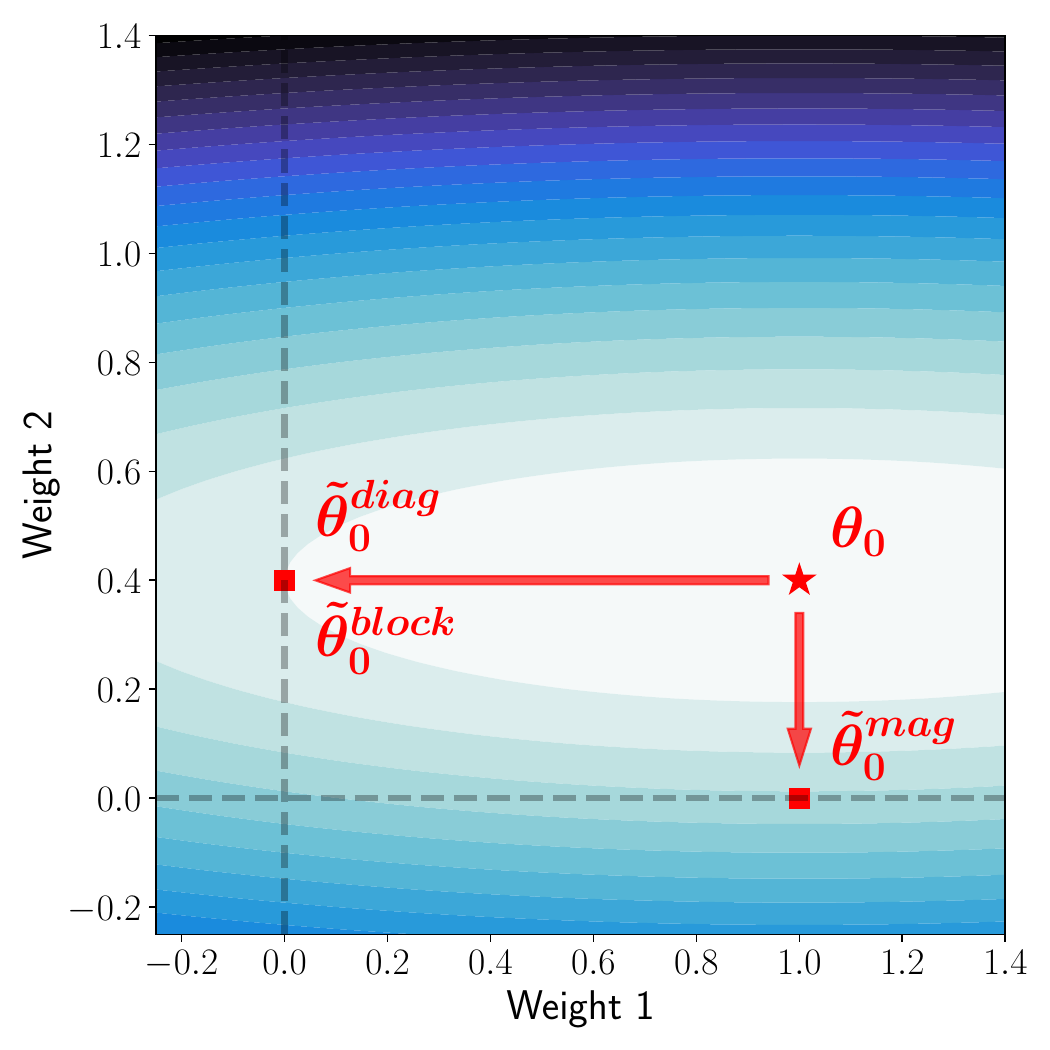}
    \includegraphics[width=0.30\linewidth]{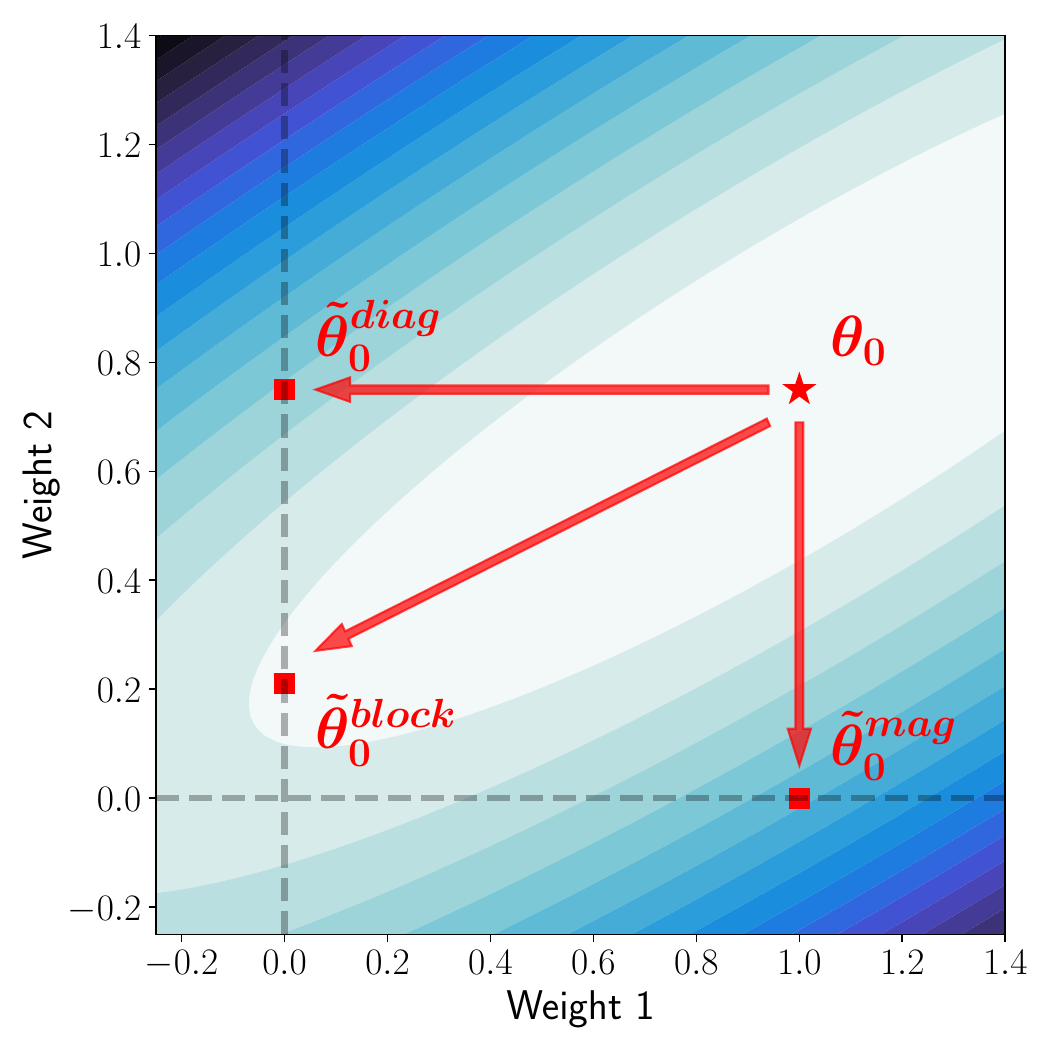}
    
    \vspace{-2.5mm}\caption{\small{\textbf{Toy loss landscapes based on different Hessian eigenstructures.} We visualize loss landscapes of toy models with two parameters, modeled as quadratic functions that compute the squared distance from the minimum $\bm\theta_0$. The curvature is defined by rescaled eigenvalues of the Hessian to simulate different task geometries. 
    \textbf{(Left)} Isotropic Hessian.
    \textbf{(Center)} Diagonal Hessian.
    \textbf{(Right)} Hessian with non-zero cross-terms.
    Pruned weight configurations, denoted $\tilde{\bm\theta}_0$, 
    are obtained using Hessian-based importance scores detailed in \cref{sec:pai_pipeline} ($\hat{\varepsilon}_{[j]}^{mag}$, $\hat{\varepsilon}_{[j]}^{diag}$, $\hat{\varepsilon}_{[j]}^{block}$). The red arrow indicates the weight perturbation, resulting from pruning. Light colors indicate lower loss.}
    }
    \label{fig:hessian_approximations}
    \vspace{-3mm}
\end{figure}

\smallskip\noindent
To intuitively understand when these three Hessian approximations might capture sufficiently well the true Hessian structure, we visualize in \cref{fig:hessian_approximations} three cases.
In the left plot, the loss landscape is isotropic so different pruning criteria lead to similar outcomes, and re-training would have little effect. In the center plot, the landscape is anisotropic, but aligned with the canonical basis. Here, magnitude pruning may be suboptimal, but re-training would not significantly help. In the right plot, the Hessian has non-zero cross-terms, creating a tilted landscape. In this case, re-training would be essential to get back to a lower-loss region.

\smallskip
\noindent
In the next section, we define our protocol for measuring how pruning affects the feature quality of pre-trained vision models, and how any resulting degradation influences transfer learning performance. 

\vspace{-4mm}
\subsection{Feature Quality Evaluation Protocol}\label{sec:feature_quality_protocol}

Inspired by \cite{kang2019featurequality}, given a task $\mathcal{D}_k$ we measure \emph{feature quality} as the ability of a classifier to linearly separate the features $\bm\Phi_k = [\varphi(\bm{x}_1, \bm{W}_0), \dots, \varphi(\bm{x}_n, \bm{W}_0)]^\top \in \mathbb{R}^{n \times h}, \forall \bm{x} \in \mathcal{D}_k$ into their correct classes. As done in \cite{bjorck2024numerical, rifkin2003regularized, fani2024fed3r}, to avoid introducing extra trainable parameters, we fit one linear classifier per task, obtaining $\bm{\xi}_{k=s,t}^\star$ by solving the following Ridge Regression \cite{boyd2004convex} problem:
\begin{equation}
    \bm{\xi}^\star_k = {\arg\min}_{\bm{\xi} \in \mathbb{R}^{h \times C}} \| Y_k - \bm\Phi_k\bm{\xi}\|_2^2 + \alpha \|\bm{\xi}\|_2^2 ~,
\end{equation}
where $Y_k \in \mathbb{R}^{n \times C}$ is the matrix of the stacked one-hot-encoding vectors of the corresponding $C$ classes and $\alpha > 0$ is an hyperparameter controlling the strength of the regularization term. This problem admits a closed-form solution:
\begin{equation}
    \bm\xi_k^\star = (\bm\Phi_k^\top\bm\Phi_k + \alpha I)^{-1} \bm\Phi_k^\top Y_k ~,
\end{equation}
where $I$ is the $h \times h$ identity matrix. As our primary focus is not on the effect of regularization, we set $\alpha = 1$ to ensure the invertibility of $(\bm\Phi_k^\top\bm\Phi_k + \alpha I)$ \cite{boyd2004convex}.

\smallskip
\noindent We keep the obtained classifiers $\bm{\xi}_s^\star, \bm{\xi}_t^\star$ frozen for all the subsequent operations and, based on which data the network is processing, we load the respective classifier. Thus, to quantify the degradation in feature quality due to pruning,

\smallskip\noindent\textbf{1. Unpruned}: we first obtain baseline predictions from the original unpruned model $\forall \bm{x} \in \mathcal{D}_k, ~f(\bm{x}, \{\bm{W}_0, \bm\xi_k^\star\}) = \varphi(\bm{x},\bm{W}_0)^\top \bm\xi_k^\star$, for all tasks $k=s,t$; 

\smallskip\noindent\textbf{2. Pruned}: we assume having access only to the \emph{source} task $\mathcal{D}_s$ and calibrate the mask $\hat{\bm{c}} \in \{0,1\}^{p - h \times C}$ for pruning. We obtain the predictions with the pruned model $\forall \bm{x} \in \mathcal{D}_k, ~f(\bm{x}, \{\bm{W}_0 \odot \hat{\bm{c}}, \bm\xi_k^\star\}) = \varphi(\bm{x},\bm{W}_0\odot \hat{\bm{c}})^\top \bm\xi_k^\star$, for all tasks $k=s,t$;

\smallskip\noindent\textbf{3. Pruned \& Fine-tuned}: we fit the remaining unpruned weights on the \emph{source} task $\mathcal{D}_s$, obtaining $\tilde{\bm{W}}_s^\star$ and evaluate the performance of the fine-tuned compressed model $\forall \bm{x} \in \mathcal{D}_k, ~f(\bm{x}, \{\tilde{\bm{W}}_s^\star, \bm\xi_k^\star\}) = \varphi(\bm{x}, \tilde{\bm{W}}_s^\star)^\top \bm\xi_k^\star$, for all tasks $k=s,t$.

\vspace{-3mm}
\section{Experiments}\label{sec:experiments}

In this section, we present the results to test our predictions \textbf{P1} and \textbf{P2}, conjectured in \cref{sec:predictions}. We describe our adopted experimental details and the results of our analysis that provide a positive answer to our hypotheses.

\smallskip
\noindent\textbf{Dataset.} We adopt ImageNet-10 \cite{huang2021unlearnable, wang2023neural}, consisting of 10 classification tasks with 10 classes each, derived as subsets of the ImageNet-1k dataset \cite{deng2009imagenet} (see \cref{tab:imagenet10_classes} for full details). As we aim to understand the degradation of feature quality induced by pruning, all procedures were performed exclusively on the test split (50 examples per class) to minimize external factors such as distribution shifts or class unbalancing that would complicate the experimental evaluation.

\begin{table}[t!]
    \vspace{-3mm}
    \begin{center}
    \small
    \setcellgapes{1.5pt}
    \makegapedcells
    \resizebox{0.98\textwidth}{!}{
    \begin{tabular}{l | cc| cc| cc| cc| cc| cc| cc| cc| cc| cc}
        \bottomrule      
        \multicolumn{21}{c}{\textbf{ImageNet-10 - Selected Multi-class Classification Tasks}} \\
        \hline\hline

        \multicolumn{1}{c}{} & \multicolumn{2}{c}{Class 1} & \multicolumn{2}{c}{Class 2} & \multicolumn{2}{c}{Class 3} & \multicolumn{2}{c}{Class 4} & \multicolumn{2}{c}{Class 5} & \multicolumn{2}{c}{Class 6} & \multicolumn{2}{c}{Class 7} & \multicolumn{2}{c}{Class 8} & \multicolumn{2}{c}{Class 9} & \multicolumn{2}{c}{Class 10} \\ \hline
        
        Task 1 (Food) & cheeseburger & (933) & carbonara & (959) & hot dog & (934) & pizza & (963) & ice cream & (928) & burrito & (965) & mashed potato & (935) & cucumber & (943) & pretzel & (932) & spaghetti & (940) \\ \hline
        
        Task 2 (Geo Formation) & sandbar &(977) & coral reef &(973) & volcano &(980) & promontory &(976) & valley &(979) & lakeside &(975) & bubble &(971) & geyser &(974) & cliff &(972) & alp &(970) \\ \hline
        
        Task 3 (Birds) & great gray owl &(24) & robin &(15) & rooster &(7) & hen &(8) & ostrich &(9) & brambling &(10) & goldfinch &(11) & bulbul &(16) & bald eagle &(22) & jay &(17) \\ \hline
        
        Task 4 (Structures) & castle & (483) & patio & (706) & bridge & (821) & obelisk & (682) & totem & (863) & dock & (536) & lighthouse & (437) & fountain & (562) & sawmill & (634) & dam & (525) \\ \hline
        
        Task 5 (Clothing) & trench coat & (869) & lab coat & (617) & poncho & (735) & bow tie & (457) & sweatshirt & (841) & kimono & (614) & jersey & (610) & miniskirt & (655) & bikini & (445) & bulletproof vest & (465) \\ \hline
        
        Task 6 (Furniture) & toilet seat & (861) & china cabinet & (495) & table lamp & (846) & desk & (526) & studio couch & (831) & stove & (827) & folding chair & (559) & pool table & (736) & bookcase & (453) & wardrobe & (894) \\ \hline
                
        Task 7 (Equipment) & cellular telephone & (487) & cd player & (485) & oscilloscope & (688) & printer & (742) & polaroid camera & (732) & computer keyboard & (508) & coffeepot & (505) & mouse & (673) & ipod & (605) & paintbrush & (696) \\ \hline
                
        Task 8 (Vehicles) & bullet train & (466) & school bus & (779) & police van & (734) & garbage truck & (569) & fire engine & (555) & pickup truck & (717) & limousine & (627) & ambulance & (407) & go-kart & (573) & tractor & (866) \\ \hline
                
        Task 9 (Land Animals) & king snake & (56) & mud turtle & (35) & zebra & (340) & lion & (291) & tiger & (292) & red fox & (277) & timber wolf & (269) & cardigan & (264) & ice bear & (296) & brown bear & (294) \\ \hline
                
        Task 10 (Aquatic Animals) & tench & (0) & goldfish & (1) & great white shark & (2) & tiger shark & (3) & hammerhead & (4) & electric ray & (5) & stingray & (6) & king crab & (121) & crayfish & (124) & jellyfish & (107) \\ \hline
        
        \bottomrule
    
    \end{tabular}}
    \end{center}
    \caption{\small We define 10 multi-class classification tasks from ImageNet-1k \cite{deng2009imagenet}, each comprising 10 semantically related classes. To ensure semantic consistency within each task, classes are grouped based on the WordNet Hierarchy \cite{miller1995wordnet}. For each task, we report the task number, the class names, and their corresponding ImageNet class indices.} 
    \vspace{-6mm}
    \label{tab:imagenet10_classes}
\end{table}

\smallskip
\noindent\textbf{Architectures \& Pre-trained Models.}
We focus on both residual networks (ResNet-50) \cite{he2016deep} and vision transformers (ViT-B/16) \cite{dosovitskiy2020image} architectures, leveraging different pre-training paradigms: supervised (on ImageNet-1k \cite{deng2009imagenet}), self-supervised contrastive (MoCoV2 \cite{chen2020improved}), self-supervised student-teacher distillation (DINO \cite{caron2021emerging}), and vision-language alignment (CLIP \cite{radford2021learning}).

\smallskip
\noindent\textbf{Baselines.} To study how pruning affects transferability, based on \cite{iurada2024finding} we resort to data-driven PaI methods: SNIP \cite{lee2018snip}, GraSP \cite{grasp2020wang} and PX \cite{iurada2024finding}, although we skip the latter on ViT, as its score applies only for networks in the NTK regime, which is problematic due to their highly non-linear self-attention layers \cite{wu2023convergence}.

\smallskip
\noindent\textbf{Implementation details.} We follow standard PaI procedures \cite{lee2018snip, tanaka2020synflow, iurada2024finding} as outlined in \cref{sec:pai_pipeline}, while the evaluation follows \cite{kang2019featurequality} as outlined in \cref{sec:feature_quality_protocol}. In detail, we first fit one classifier $\bm\xi_k^\star$ per task $k=1,\dots,10$ on the unpruned model and keep them frozen at all times. 
Then, at each run one task is elected as \emph{source} while the other nine are held-out as \emph{transfer} tasks. We prune the pre-trained encoder $\bm{W}_0$ by calibrating the pruning mask $\bm{c}$, accessing only the \emph{source} task $\mathcal{D}_s$ when needed. Then, we re-train the pruned encoder on $\mathcal{D}_s$, aligning with \cite{iurada2024finding}. 
We run SGD on MSE loss for 90 epochs with 0.9 momentum and $10^{-4}$ weight decay. The step size is $5\times 10^{-5}$ (except for ViT DINO, where $10^{-4}$ was used) and it is decayed by a factor $\times 10$ at epochs 30, 60, 80. For ResNet models we avoid updating the normalization statistics. 
Finally, akin to \cite{ntksap2023wang}, we evaluate each method at \{36.00, 47.52, 59.04, 66.42, 73.80, 78.52, 83.22\} sparsities (\%).

\begin{figure}[t!]
    \centering
    \vspace{-3mm}

    \includegraphics[width=0.31\linewidth]{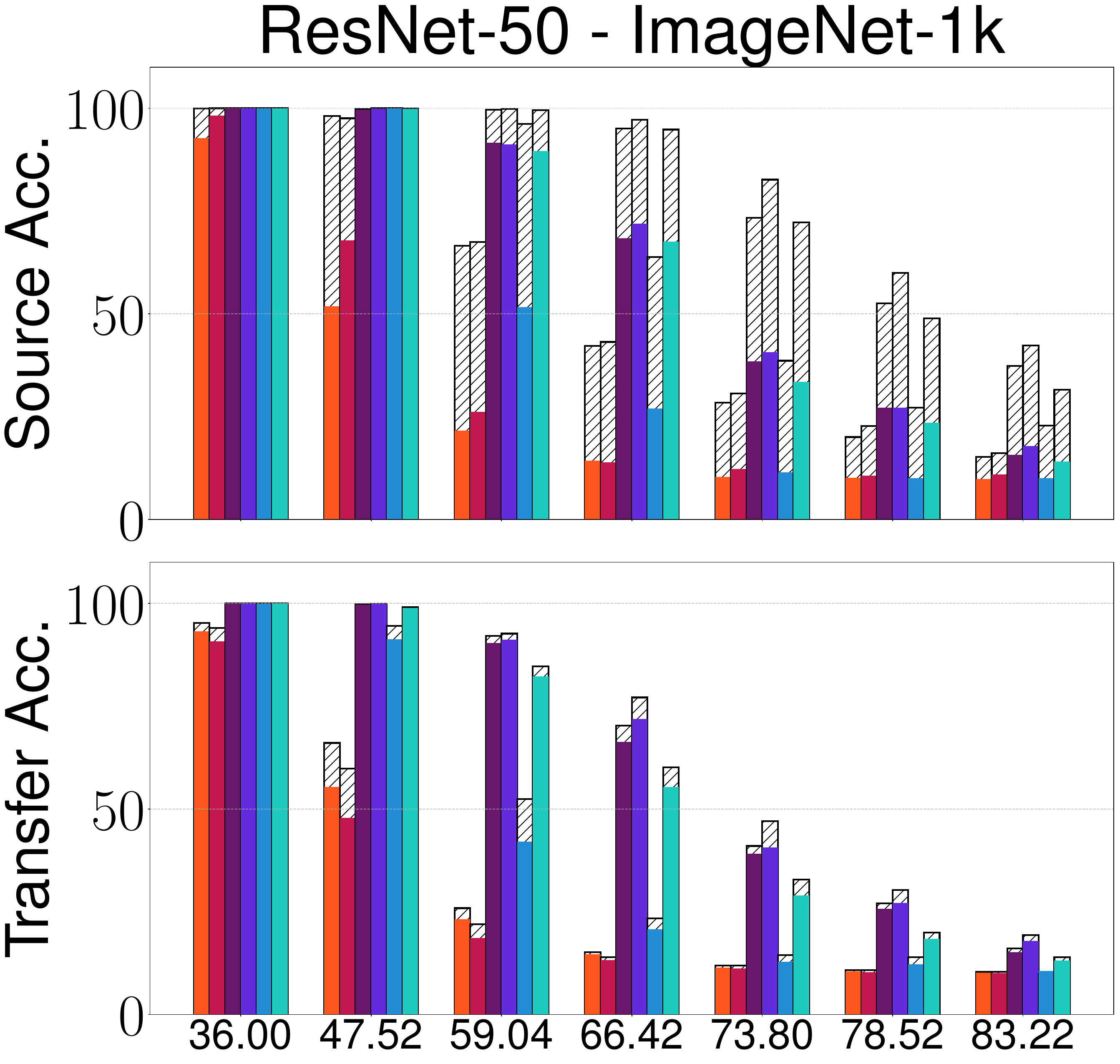}
    \includegraphics[width=0.31\linewidth]{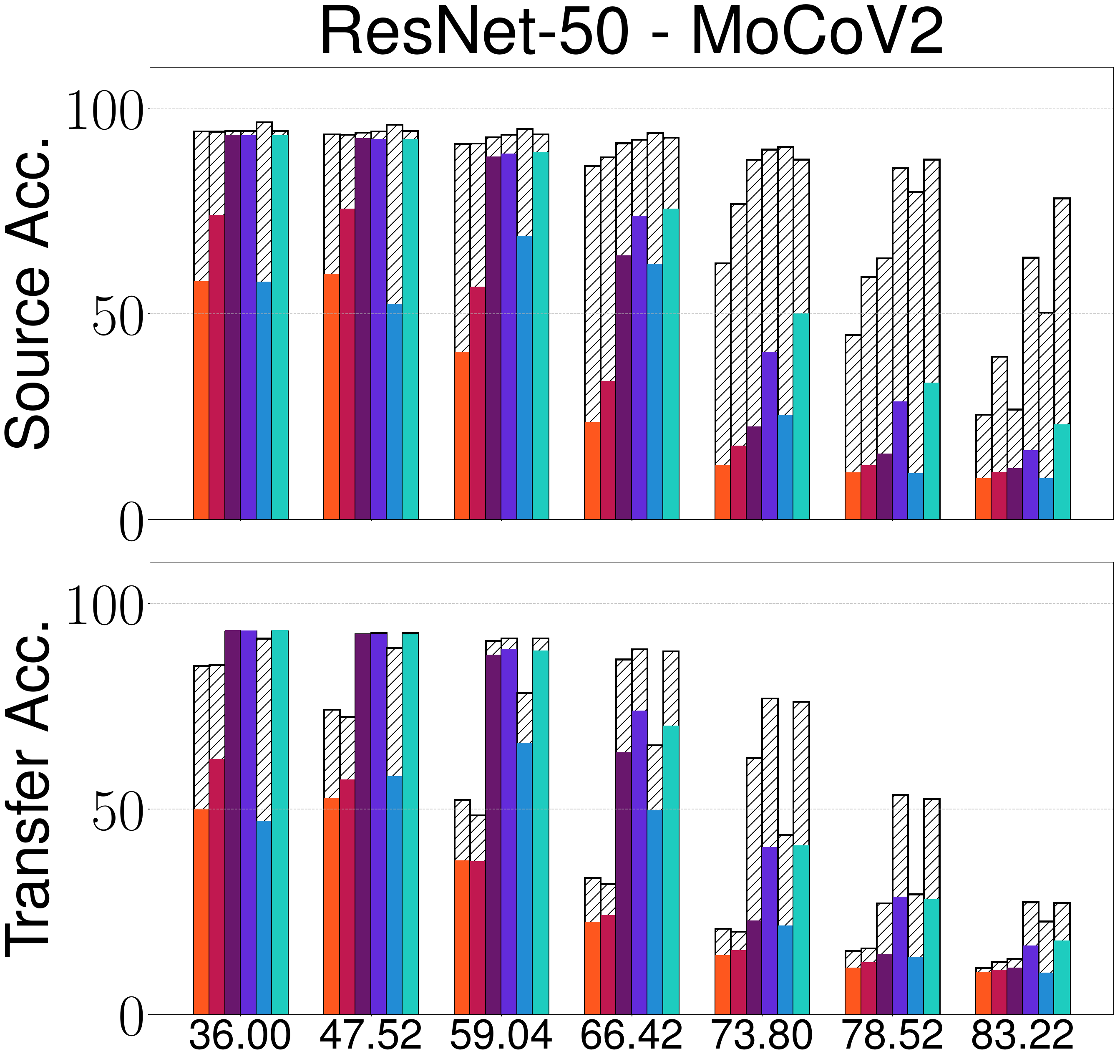}
    \includegraphics[width=0.31\linewidth]{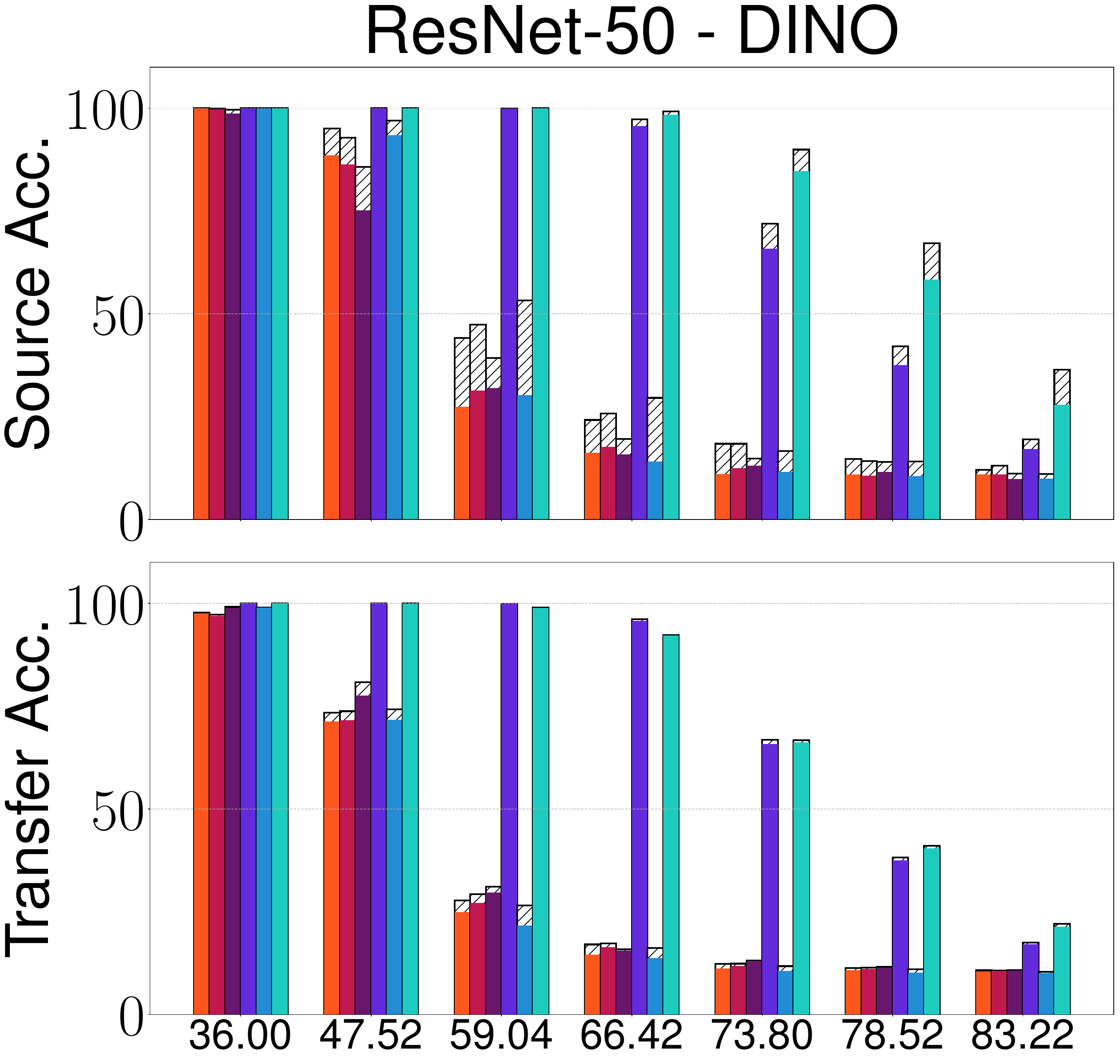}

    \vspace{3mm}
    \includegraphics[width=0.31\linewidth]{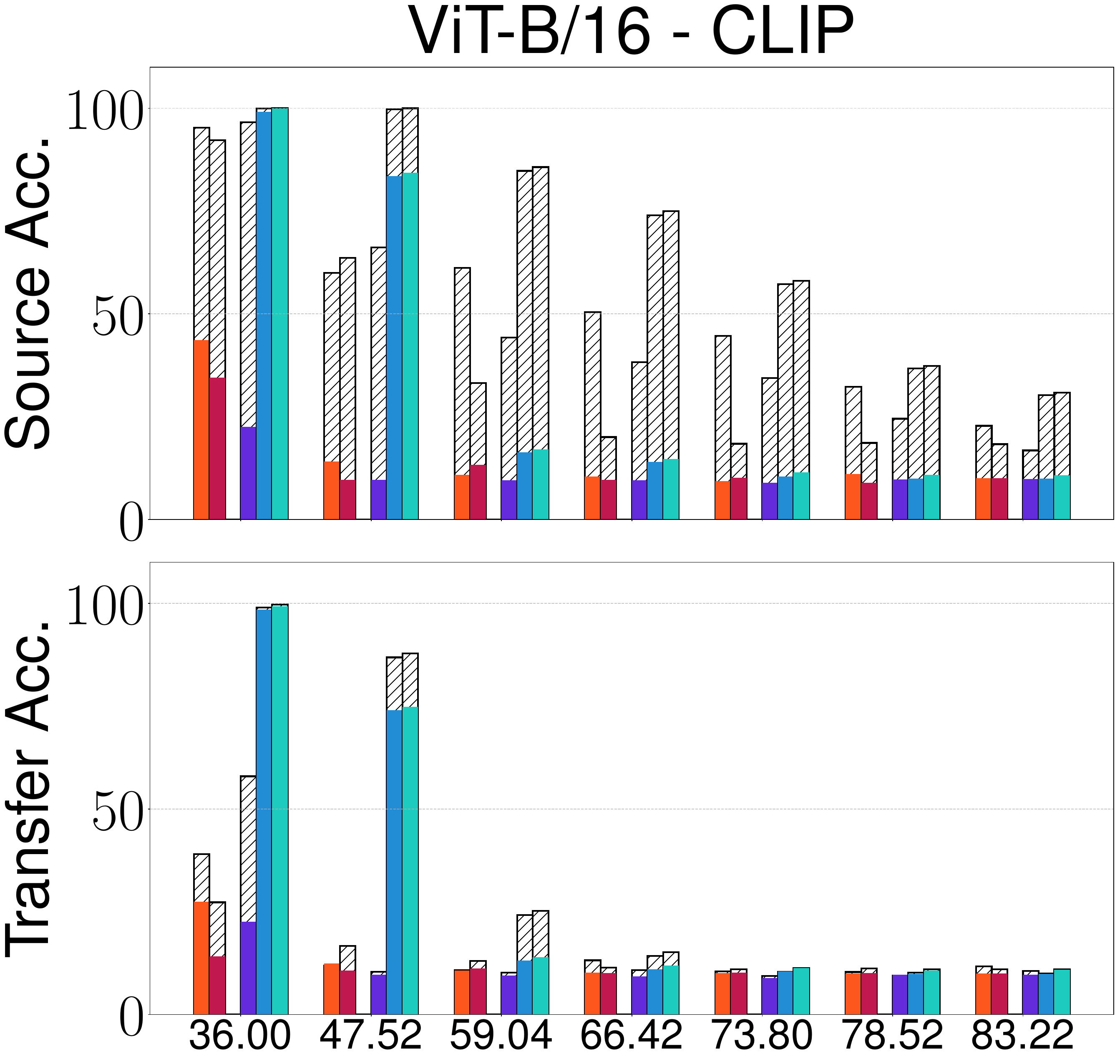}
    \hspace{8pt}
    \includegraphics[width=0.31\linewidth]{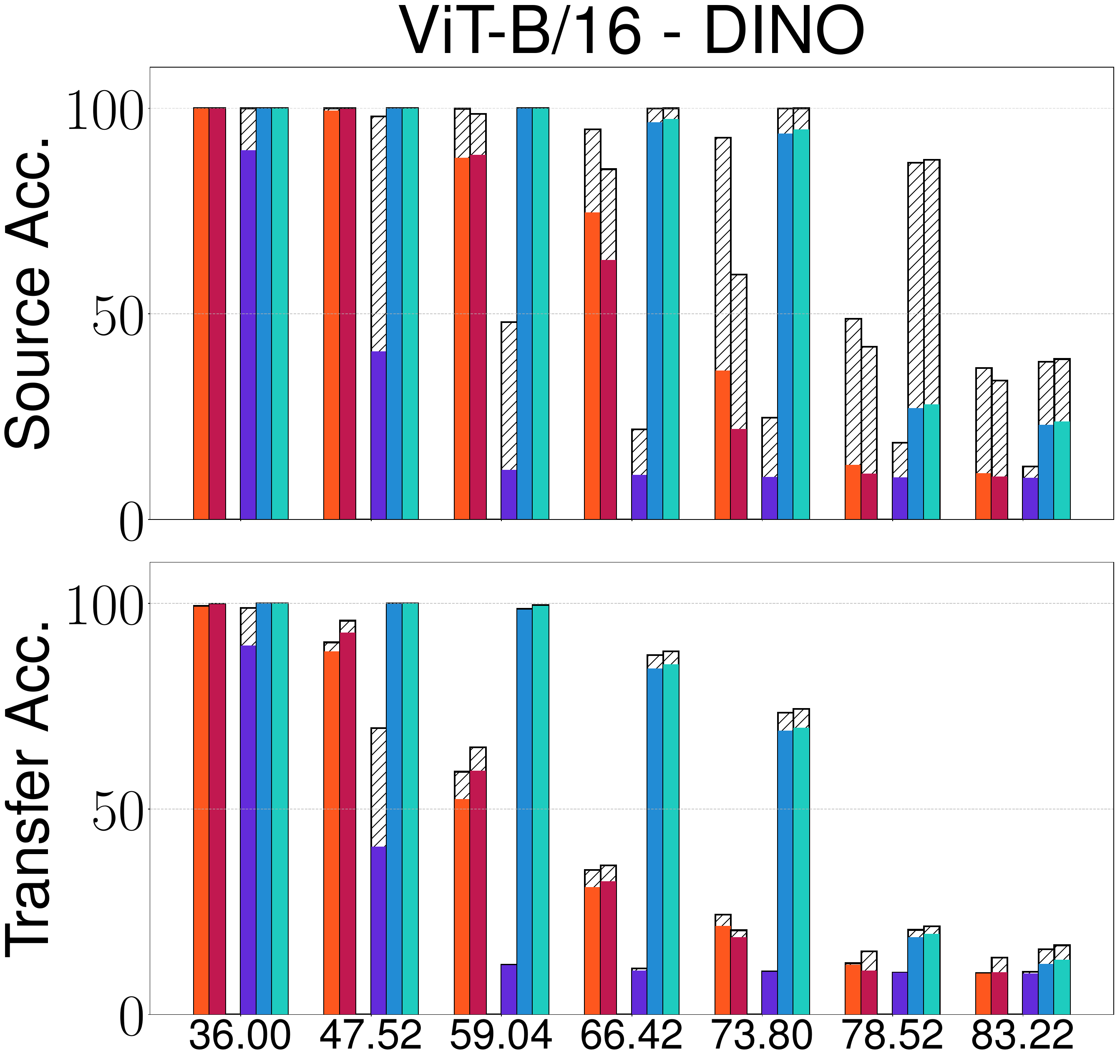}

    \vspace{1mm}
    \includegraphics[trim=20 40 10 40, clip, width=0.6\linewidth]{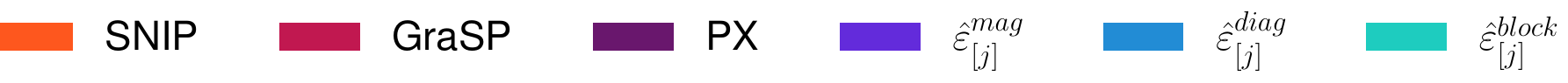}
    
    \vspace{-2.5mm}\caption{\small\textbf{Source and transfer performance of pruned models.} We report the average accuracy of models pruned at increasing sparsity ratios using different scores, where pruning is performed solely based on the \emph{source} task. Solid bars indicate zero-shot accuracy on the pruned model.
    Hatched bars represent the accuracy after re-training the pruned model on the \emph{source} task; in all cases, re-training leads to improved performance. For each model, we show the average accuracy on the \emph{source} task (top) and the average accuracy across the \emph{transfer} tasks (bottom). Accuracy is averaged over 10 runs, where in each run a different task is designated as the \emph{source}, and the remaining nine are treated as \emph{transfer} tasks.
    }
    \label{fig:hist_performance}
    \vspace{-3mm}
\end{figure}

\vspace{-4mm}

\subsection{Impact of Source‐Only Pruning on \emph{Source} and \emph{Transfer} Tasks}

In \cref{fig:hist_performance}, each cluster along the horizontal axis corresponds to a target sparsity level. Within each cluster, the six colored bars represent different pruning scores. Solid bars show zero‐shot accuracy of the pruned model; black‐hatched bars indicate accuracy after fine‐tuning on the \emph{source} task. 

\noindent\textbf{Pruning on \emph{source} preserves \emph{transfer} (P1).}
Across all ResNet models (ImageNet-1k, MoCoV2, DINO), magnitude ($\hat\varepsilon^{mag}_{[j]}$) and block Hessian ($\hat\varepsilon^{block}_{[j]}$) pruning maintain both source and transfer performance up to high sparsities, with parallel trends between solid bars in the top and bottom rows, thus supporting prediction \textbf{P1}: these criteria prune largely redundant weights. In contrast, diagonal Hessian ($\hat\varepsilon^{diag}_{[j]}$) pruning, SNIP, GraSP, and PX show sharper declines, indicating the removal of critical parameters.

\noindent A similar yet distinct pattern holds for ViTs: DINO models retain transfer accuracy under diagonal and block Hessian pruning, whereas CLIP models degrade sooner, suggesting lower compressibility. For ViTs, diagonal Hessian is the only score besides block Hessian that reliably preserves source and transfer performance, again supporting \textbf{P1}.

\noindent\textbf{Fine‐tuning on \emph{source} helps (and never hurts) \emph{transfer} (P2).}
Examining the hatched bars in \cref{fig:hist_performance}, fine-tuning on the \emph{source} task never reduces transfer accuracy; often it improves it. This is especially notable for MoCoV2, which has greater headroom for transfer gains. In DINO-pretrained models, gains are smaller, reflecting their already strong zero-shot transfer performance, thus, supporting \textbf{P2}: brief source-task fine-tuning enhances or preserves transfer ability.
Compared to Hessian-based scores, which approximate the optimal pruning conditions, SNIP, GraSP and PX exhibit significantly steeper drops in both source and transfer accuracy at moderate sparsity.

\begin{figure}[t!]
    \centering
    \vspace{-3mm}
    \setlength\fboxsep{0pt} 
    \setlength\fboxrule{0pt}

    \fbox{%
        \begin{minipage}{0.43\linewidth}
            \centering
            \scriptsize{ResNet-50 - MoCoV2} \\
            \includegraphics[trim=0 0 60 0, clip, width=0.40\linewidth]{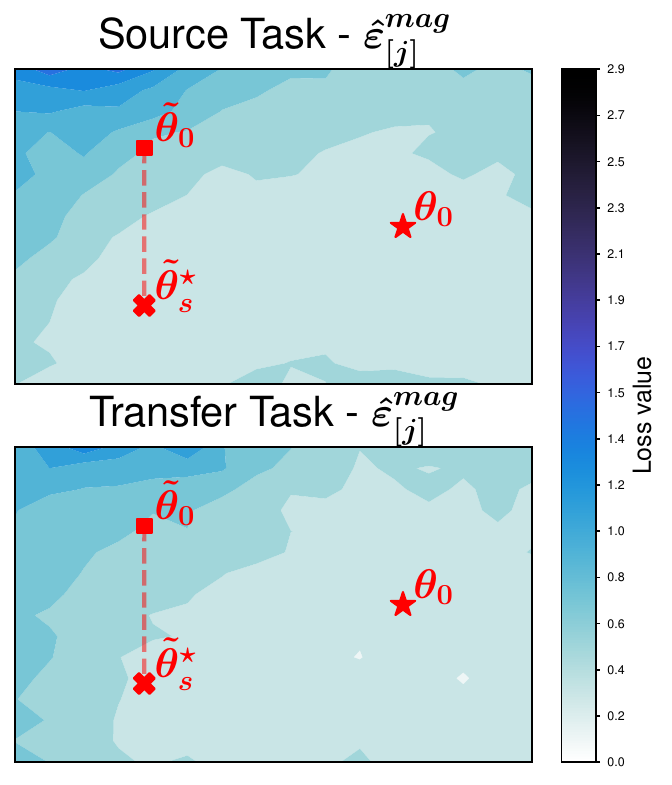}
            \includegraphics[width=0.49\linewidth]{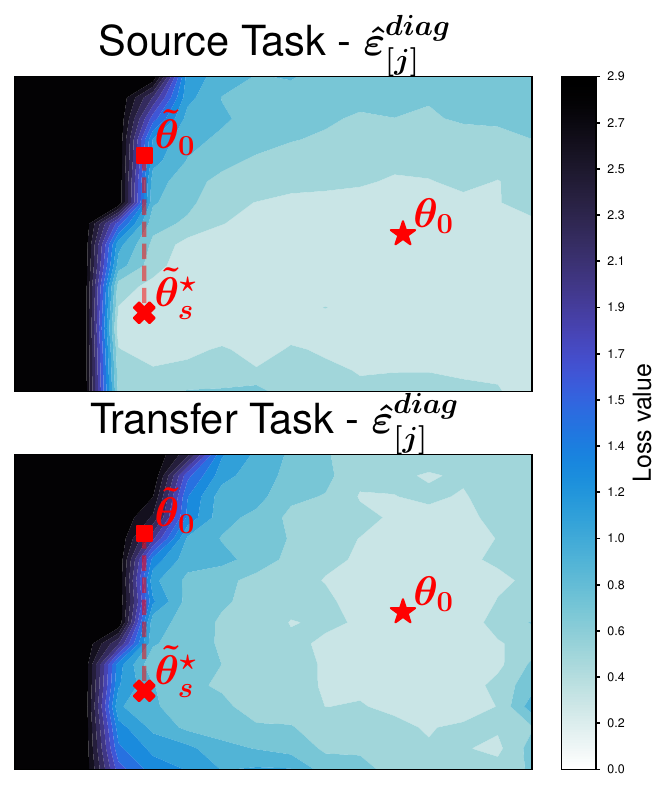}
        \end{minipage}
    }
    \hspace{1mm}
    \fbox{%
        \begin{minipage}{0.43\linewidth}
            \centering
            \scriptsize{ResNet-50 - DINO} \\
            \includegraphics[trim=0 0 60 0, clip, width=0.40\linewidth]{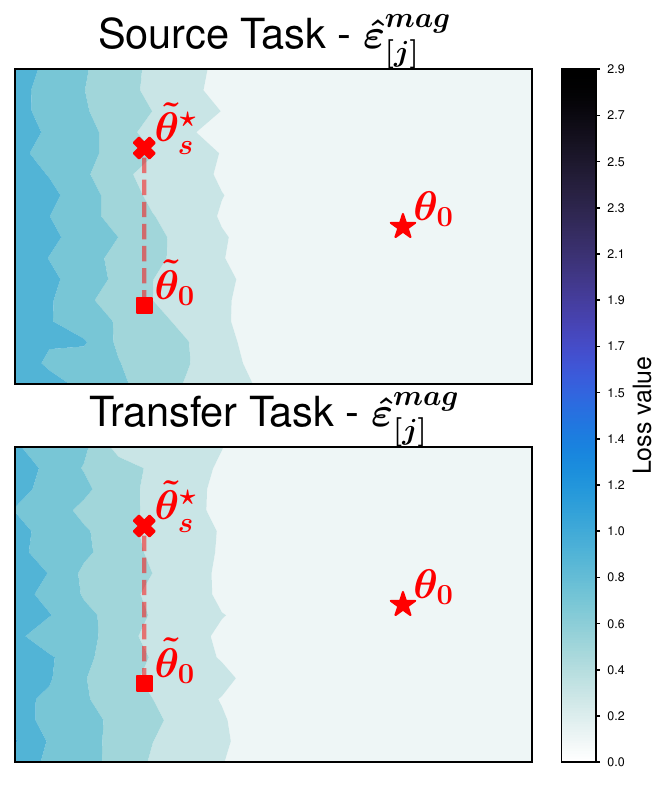}
            \includegraphics[width=0.49\linewidth]{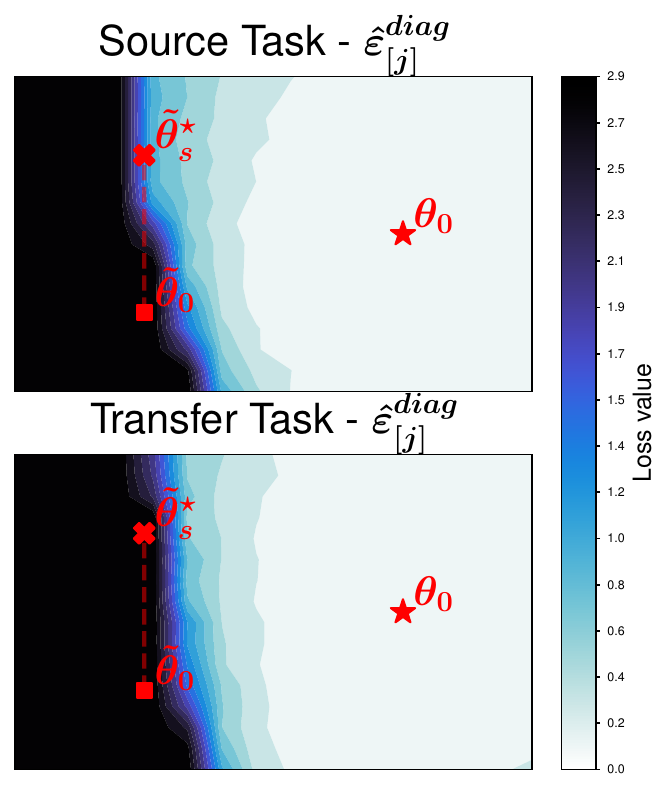}
        \end{minipage}
    }

    \vspace{1mm}

    \fbox{%
        \begin{minipage}{0.43\linewidth}
            \centering
            \scriptsize{ViT-B/16 - CLIP} \\
            \includegraphics[trim=0 0 60 0, clip, width=0.40\linewidth]{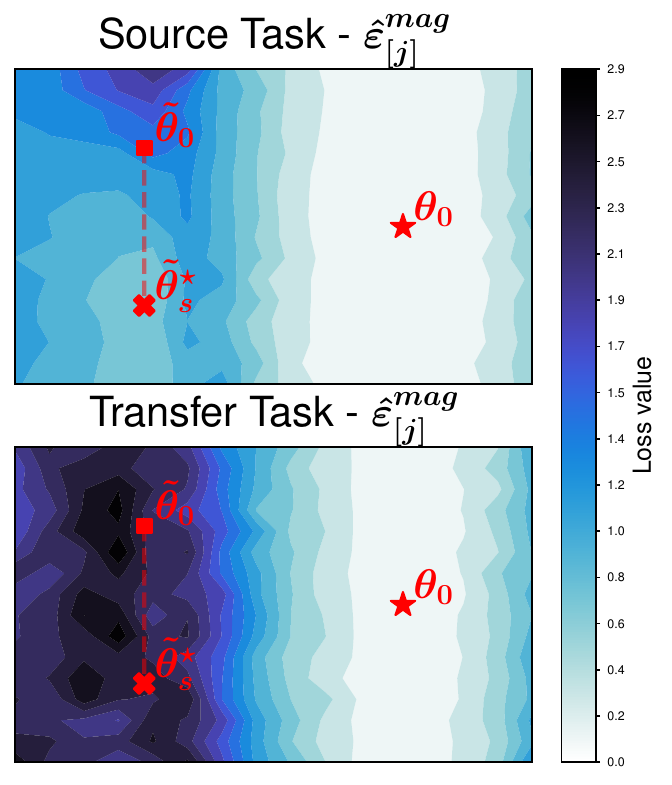}
            \includegraphics[width=0.49\linewidth]{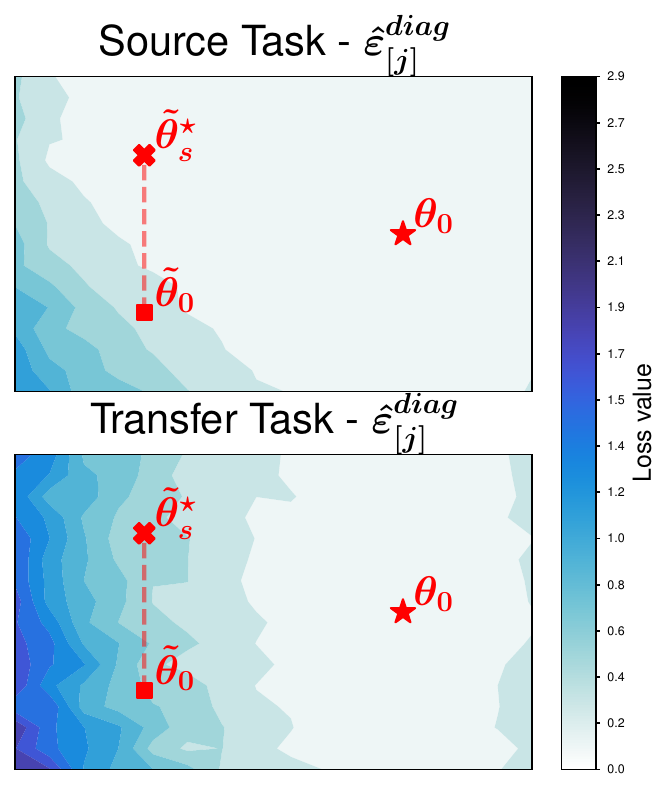}
        \end{minipage}
    }
    \hspace{1mm}
    \fbox{%
        \begin{minipage}{0.43\linewidth}
            \centering
            \scriptsize{ViT-B/16 - DINO} \\
            \includegraphics[trim=0 0 60 0, clip, width=0.40\linewidth]{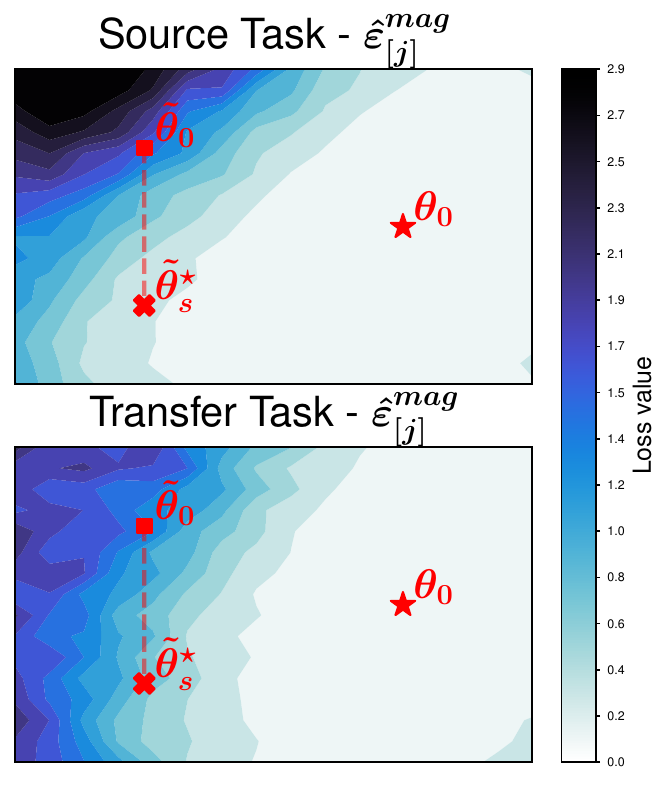}
            \includegraphics[width=0.49\linewidth]{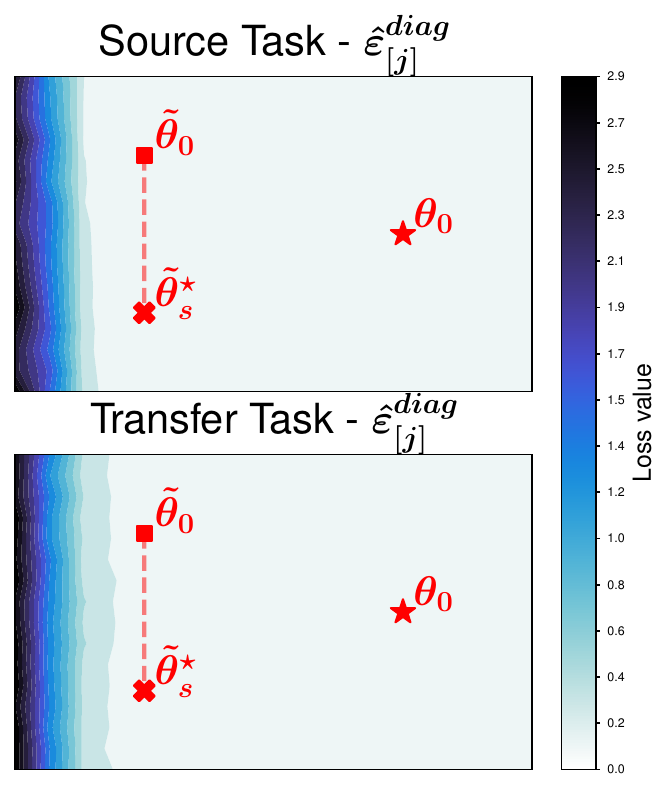}
        \end{minipage}
    }

    \vspace{-2.5mm}
    \caption{\small\textbf{Loss landscapes of real pre-trained models.} We visualize the effect of pruning and re-training on Task 1 (\emph{source}) and compare it to the corresponding behavior on Task 10 (\emph{transfer}), using the pruning scores $\hat\varepsilon_{[j]}^{mag}$ and $\hat\varepsilon_{[j]}^{diag}$. Each plot is a projection onto the subspace spanned by the top two principal eigenvectors of the covariance matrix, computed from the concatenated weights $[\bm\theta_0, \tilde{\bm\theta}_0, \tilde{\bm\theta}_s^\star]$: where $\bm\theta_0$ are the original pre-trained weights, $\tilde{\bm\theta}_0$ are the weights after pruning (on the source task), and $\tilde{\bm\theta}_s^\star$ are the weights after both pruning and re-training (also on the source task). ResNet models are pruned to 66.42\% sparsity, and ViT models to 47.52\%.
    }
    \label{fig:loss_landscapes}
    \vspace{-3mm}
\end{figure}

\smallskip\noindent
To further test our findings and provide a more interpretable perspective, we visualize the loss landscapes of two tasks, one designated as the \emph{source} and the other as the \emph{transfer} task, in \cref{fig:loss_landscapes}. We consistently choose tasks 1 and 10 for this analysis, as we observed similar patterns across other task pairs.
Focusing on ResNet-50 models, we find that magnitude pruning ($\hat\varepsilon_{[j]}^{mag}$) results in more favorable loss landscapes. This observation supports prediction \textbf{P1} for ResNet-50. Moreover, we see that fine-tuning tends to move the solution toward lower-loss regions not only for the \emph{source} task but also for the \emph{transfer} task, particularly in the case of ResNet-50 MoCoV2. In contrast, this trend does not hold as clearly for ResNet-50 DINO, aligning with prediction \textbf{P2}. On the other hand, diagonal Hessian-based pruning ($\hat\varepsilon_{[j]}^{diag}$) appears less effective, as it tends to prune important directions in weight space and results in less favorable loss landscapes. Still, when re-training improves performance on the source task, it generally helps move toward better solutions for the transfer task as well.
Switching to ViT models, we observe the opposite: diagonal Hessian-based pruning performs better than magnitude-based pruning, though the overall trends remain consistent. Notably, for ViT-B/16 DINO, the lack of dramatic differences seen in \cref{fig:hist_performance} can be attributed to already high zero-shot performance on both source and transfer tasks, limiting the observable gains.
\vspace{-4mm}
\section{Conclusion}

\vspace{-3mm}

Our results challenge the prevailing view that pruning without task-specific guidance necessarily harms transferability. We show that vision models retain zero-shot performance on \emph{transfer} tasks even when compressed on \emph{source} task data. Moreover, fine-tuning these pruned models often enhances \emph{transfer} performance, suggesting a surprising alignment in the loss landscapes across tasks.
This robustness holds across architectures, though ViTs require more sophisticated Hessian-based criteria, while ResNets perform best with simpler magnitude pruning. 

\noindent
Nonetheless, our controlled setup omits real-world complexities. Also, theoretical guarantees for safe pruning thresholds remain unexplored. Future research should address these gaps by linking pruning-induced perturbations to the geometry of the loss landscape and testing PaI under dynamic or heterogeneous conditions.
In summary, we show that the rich representations learned during large-scale pre-training provide a robust foundation for efficient compression, enabling lighter, faster models that retain their capacity to adapt \emph{almost} for free.

\smallskip
\footnotesize{
\noindent\textbf{Acknowledgements.}
L.I. acknowledges the grant received from the European Union Next-GenerationEU (Piano Nazionale di Ripresa E Resilienza (PNRR)) DM 351 on Trustworthy AI. T.T. acknowledges the EU project ELSA - European Lighthouse on Secure and Safe AI.
This study was carried out within the FAIR - Future Artificial Intelligence Research and received funding from the European Union Next-GenerationEU (PIANO NAZIONALE DI RIPRESA E RESILIENZA (PNRR) – MISSIONE 4 COMPONENTE 2, INVESTIMENTO 1.3 – D.D. 1555 11/10/2022, PE00000013). 
This manuscript reflects only the authors’ views and opinions, neither the European Union nor the European Commission can be considered responsible for them. We acknowledge the CINECA award under the ISCRA initiative for the availability of high-performance computing resources and support.
}
\vspace{-4mm}

%
%
\bibliographystyle{splncs04}
\bibliography{main}

\end{document}